\documentclass{article}
\usepackage{ijcai22}

\usepackage{times}
\usepackage{soul}
\usepackage{url}
\usepackage[hidelinks]{hyperref}
\usepackage[utf8]{inputenc}
\usepackage[small]{caption}
\usepackage{graphicx}
\usepackage{amsmath}
\usepackage{amsthm}
\usepackage{booktabs}
\usepackage{algorithm}
\usepackage{algorithmic}
\usepackage{multirow}
\usepackage{svg}
\usepackage{tabularx}
\usepackage{pifont} 
\newcommand{\cmark}{\ding{51}}
\newcommand{\xmark}{\ding{55}}

\newcolumntype{C}{>{\centering\arraybackslash}X}

\title{Scatter Points in Space: 3D Detection from Multi-view Monocular Images}

\author{
Jianlin Liu$^1$
\and
Zhuofei Huang$^2$\and
Dihe Huang$^3$\and
Shang Xu$^1$\and
Ying Chen$^1$\And
Yong Liu$^1$
\affiliations
$^1$Tencent\\
$^2$HKUST\\
$^3$Tsinghua University
\emails
\{jenningsliu, shangxu, mumuychen, choasliu\}@tencent.com,
zhuangbr@cse.ust.hk,
hdh20@mails.tsinghua.edu.cn
}

\begin{document}

\maketitle

\begin{abstract}
  3D object detection from monocular image(s) is a challenging and long-standing problem of computer vision. To combine information from different perspectives without troublesome 2D instance tracking, recent methods tend to aggregate multi-view feature by sampling regular 3D grid densely in space, which is inefficient. In this paper, we attempt to improve multi-view feature aggregation by proposing a learnable keypoints sampling method, which scatters pseudo surface points in 3D space, in order to keep data sparsity. The scattered points augmented by multi-view geometric constraints and visual features are then employed to infer objects location and shape in the scene. To make up the limitations of single frame and model multi-view geometry explicitly, we further propose a surface filter module for noise suppression. Experimental results show that our method achieves significantly better performance than previous works in terms of 3D detection (more than 0.1 AP improvement on some categories of ScanNet). The code will be publicly available.
\end{abstract}

\begin{figure}[t]
    \centering
    \includegraphics[width=\columnwidth]{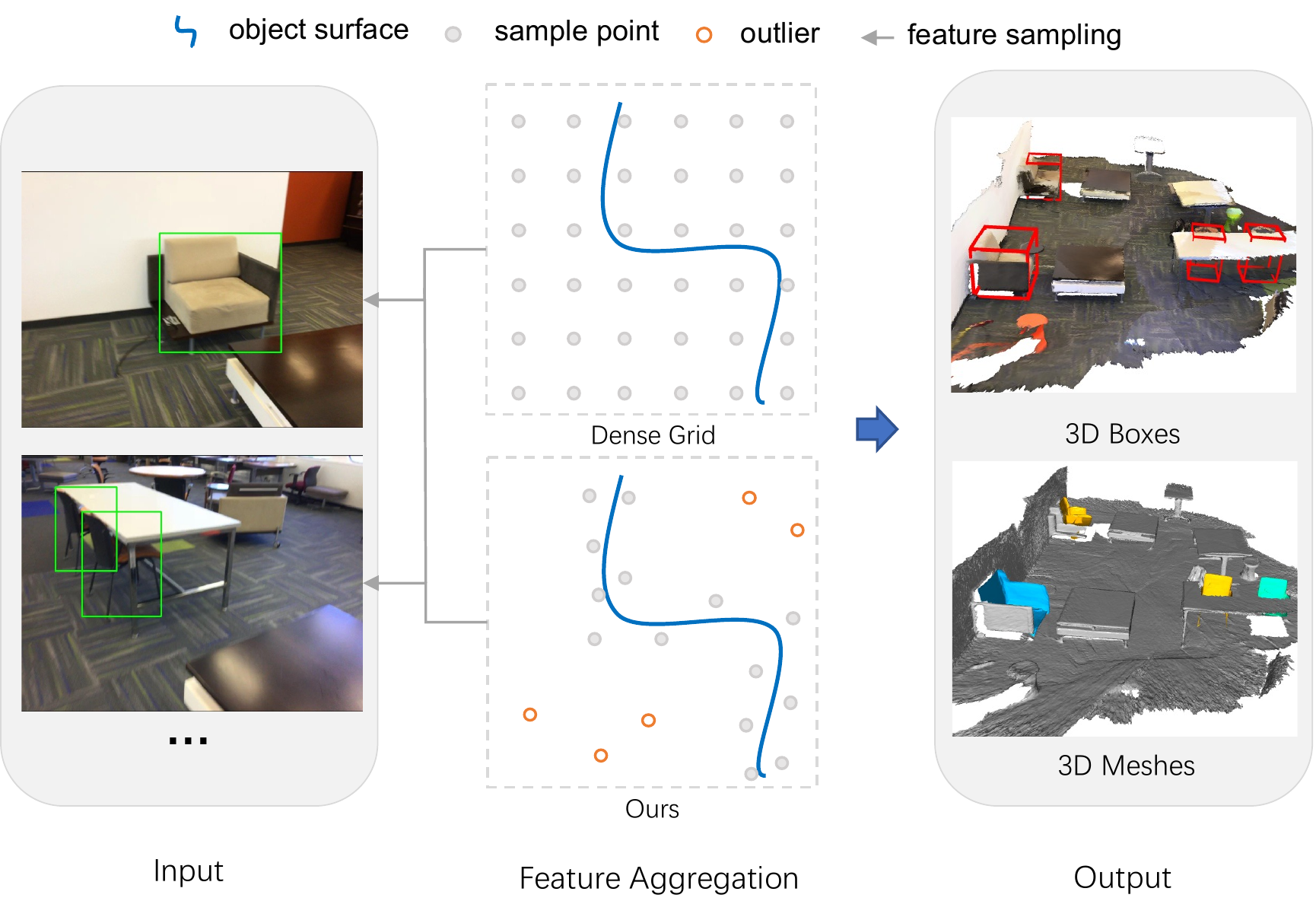}
    \caption{Problem definition. The aim is to infer 3D orientated boxes and meshes from multi-view posed RGB images with 2D detection boxes. For multi-view feature aggregation, dense grid points are prevalently used by existing methods. Our method learns to generate dynamic keypoints around object surface first as container for feature aggregation.} 
    \label{fig:fig1}
\end{figure}

\section{Introduction}

3D object detection usually requires accurate active depth sensing techniques including LiDAR and structured-light cameras, which are prohibitively expensive or limited to certain circumstances. Therefore, some researchers turn to using RGB image(s) instead. Remarkable progress has been made in the last few years for monocular 3D detection, yet the performance is limited by two major problems. On the one hand, field-of-view(FoV) is often too small to capture large object. On the other hand, estimating depth from a single monocular image is an ill-posed problem. On the contrary, leveraging multi-view images for 3D detection can help resolve these limitations. To explicitly make use of multi-view visual clue, it usually requires matching instance boxes or pixels across 2D images, which is called data association. However, data association can be really challenging when there are large transformation between image pair. Therefore, several recent works tried to perform monocular 3D objects detection/reconstruction in 3D space, which exhibits a trend of using dense 3D grid as anchor points for aggregating multi-view deep feature. Compared to aggregating multi-view features by 2D data association, accumulating temporal vision features via 3D anchor points helps better preserve 3D geometry and doesn't require any explicit data association. Noticed that dense 3D grid is used because no prior about scene structure is available. Although the idea is natural and effective, it overlooks the sparsity of scene structure. To ensure a good coverage of object surface, grid resolution must be high enough which leads to excessively large memory footprint. We claim that using continuous sampled points in 3D space to replace regular sampled 3D grid points is more efficient and effective. To generate continuous sampled anchor points without 3D structure prior, we propose a learnable 3D anchor points sampling method by back-projecting multi-view depth estimation. These continuous anchor points will be scattered around object surface, instead of evenly placed in the whole scene space.

In this paper, we focus on solving the problem of 3D detection with multiple posed RGB images. As mentioned above, we will demonstrate that pseudo 3D point cloud from the predicted depth maps is a good proxy for multi-view feature aggregation. This design not only saves memory but also boosts the performance of  downstream tasks. Specifically, the procedure of multi-frame depth maps back-projection and point cloud sampling is termed as \textbf{Point Scattering}. With camera intrinsics and poses given, the scattered points will be further augmented by fetching multi-view 2D features. Additionally, in order to remedy the inaccuracy of single-view depth prediction with multi-view geometry, we propose a surface filtering module to filter out error-prone scattered points. Based on these augmented points with refined features, we adopt sparse voxel convolution to extract both geometry and visual aware features. Various downstream tasks such as detection or reconstruction can be applied on top of the extracted 3D features. In this work, we build a VoteNet \cite{qi2019deep} based 3D detector with reconstruction head to demonstrate the effectiveness of our method. 

In summary, the key contributions of this paper are as follows:

\begin{enumerate}
    \item We propose to aggregate multi-view feature by scatter continuous points in space, which reveal rough scene skeleton. Furthermore, the scattered points is refined by a surface filtering module to get cleaner point cloud representation.
    \item We propose a novel 3D detection method based on multi-view feature augmented scattered points, which outperforms state-of-the-art multi-view monocular 3D detection methods by a large margin.
\end{enumerate}

\section{Related Works}


\paragraph{Point cloud 3D Object Detection.} Normally, 3D detection takes single frame point cloud or depth map as input. Our work focus on 3D understanding given only a series of RGB images, which is much more challenging than point cloud object detection. \cite{yan2018second} proposes to use sparse convolution to reduce computation and space complexity in point cloud detection, which shows the significance of keeping sparsity. Compared to other RGB fusion methods, \cite{vora2020pointpainting} reports superior detection result of appending 2D categorical information in point-wise manner. Different from these works, our method fuse multi-frames RGB feature with sparse pseudo point cloud, and leverage multi-view stereo to enhance geometry feature.

\paragraph{Monocular 3D Object Detection.} Single-view 3D object detection presents a challenging task in computer vision due to inaccurate depth prediction. \cite{Reading2021} tries to weight image feature in the frustum grid by predicting discrete depth distribution, and perform 3D detection on bird’s-eye-view (BEV) feature map. Another voxel-based method \cite{Liu2021} proposes to learn a regular grid of 3D voxel features from the input image which is aligned with 3D scene space via a 3D feature lifting operator. Based on the 3D voxel features, its CenterNet-3D detection head formulates the 3D detection as keypoint detection in the 3D space. \cite{Nie2020} assumes each in-room object has a multi-lateral relation between its surroundings, and takes all of them into account in predicting its bounding box. However, most of monocular schemes suffer from the small Field-Of-View(FoV) when capturing the large object (eg. large dining table or conference table). Besides, above voxel-based methods \cite{Reading2021} \cite{Liu2021} may easily cause waste of memory while building voxel grids, since very few voxels are occupied by objects. In this work we propose a more memory efficient point-based framework, which only back-project some keypoints into 3D space.

\paragraph{Multi-view 3D Object Detection.} How to combine information from multiple frames is one major problem of multi-view 3D understanding. To solve this, some works proposed to detect objects in each frame separately, and combine results across frames by data association. In \cite{runz2020frodo} a 3D ray clustering strategy is utilized to match 2D detection results, which may fail when objects are close to each other. From the associated multi-view 2D boxes, \cite{runz2020frodo} use \cite{nicholson2018quadricslam} and post optimization to reconstruct 3D objects. MOLTR \cite{li2021moltr} is an unified framework for object-centric mapping from RGB videos, which consists of monocular 3D detection, multiple model Bayesian filter tracking and shape code prediction. However, same as \cite{Luiten2019}, data association is achieved by tracking, continuous video frames are preferred to avoid tracking failure. More recently, researchers found that accumulating multi-view feature upon a dense 3D grid can be used to perform end-to-end 3D reconstruction. \cite{murez2020atlas} is the seminal work that accumulates image feature in 3D volume along pixel rays, which enable 3D reconstruction from multi-view images. Similarly, \cite{bozic2021transformerfusion} process monocular video by a transformer network that fuse observations into a volumetric scene representation. They are both limited by cubic space complexity caused by the dense grid representation. Our method avoids error-prone data association and is more scalable since we sample continuous point around surface to keep data sparsity.

\section{Methodology}

\begin{figure*}[htbp]
\centering
\includegraphics[width=\textwidth]{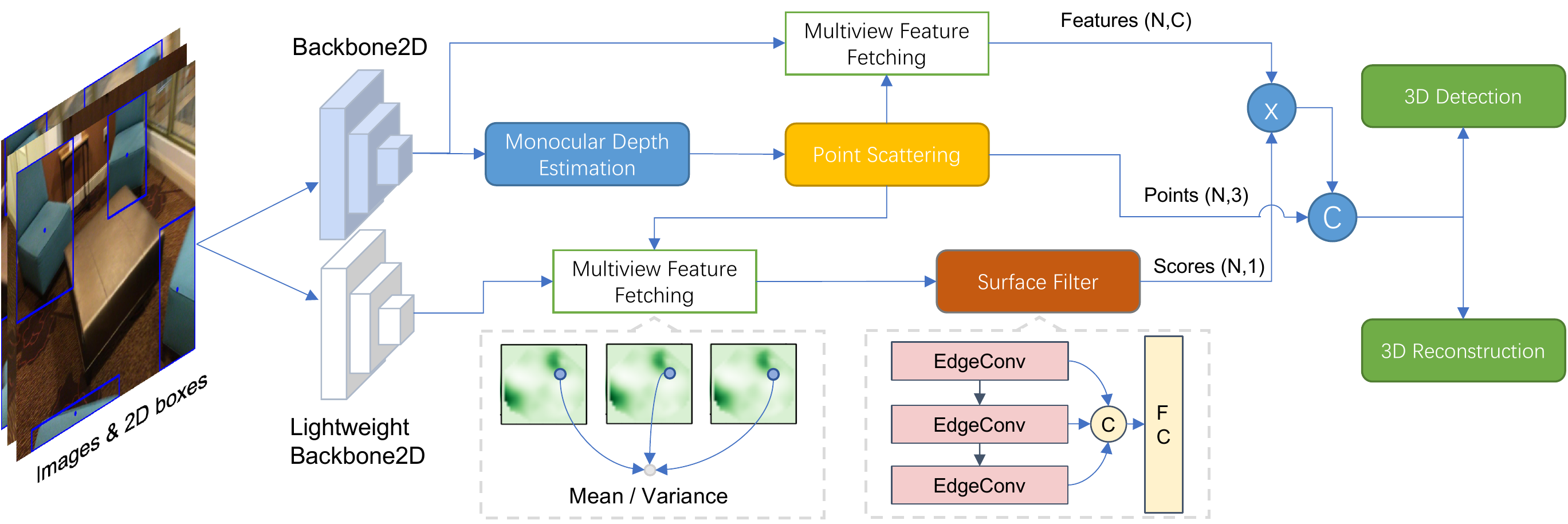}
\caption{Overview. Our method predicts depth map for each RGB frame and back-project pixels within 2D boxes to 3D space(\textbf{Multi-view Point Scattering}). After an approximately even sampling of the scattered points, 2D feature from different views are aggregated in point wise manner to facilitate downstream task. In addition, a \textbf{Surface Filter} module is introduced to down-weight the error-prone scattered points.}
\label{fig:overview}
\end{figure*}

Our method mainly consists of three modules that we will describe in sequence: monocular depth estimation, point scattering with multi-view augmentation, and point-based 3D detection. Monocular depth estimation aims to provide rough observation of partial scene. As a key contribution of this paper, continuous anchor points are generated with depth maps from different view-points and further augmented by multi-view feature. The augmented points contains geometry information captured by coordinates and feature variance across frames, as well as visual clues from multiple image frames. Taking these augmented points as cornerstone, downstream 3D understanding tasks can be applied, such as 3D detection and reconstruction. In this paper, we build a VoteNet\cite{qi2019deep} based 3D detection and reconstruction module by voxelizing the aforementioned scattered points with augmented features. 
Our method takes RGB images with known camera poses and instrisic parameters, as well as optional 2D detection results. 3D information is reconstructed by monocular depth estimation and multi-view stereo instead of active sensing. How to get 2D detection is beyond discussion of this paper, which can be easily obtained by any off-the-shelf 2D detectors such as \cite{ren2016faster} in practice. To isolate the performance of 2D detector, we generate 2D boxes by projecting 3D ground-truth meshes to each frame. As described in \ref{para:keypoints_generation}, 2D detection is used in two aspects, (1) reduce pixels being converted to 3D points. (2) append categorical information to the scattered points. 

\subsection{Monocular Depth Estimation}

Without 3D prior of the scene, it is harder to build anchor points other than volume grid points. Therefore, we resolve to choose monocular depth estimation as a proxy for learning 3D anchor points. For depth estimation, we use ordinal regression instead of direct regression following the advanced monocular depth estimation method \cite{fu2018deep}. We simplify the network structure in \cite{fu2018deep} to reduce the amount of parameters since it works as a submodule in our end-to-end framework. We remove the Scene Understanding Layer the decoder from \cite{fu2018deep}, and downsample the output resolution of depth map by factor 4. As the depth range of our major dataset ScanNet is relatively small, we apply linear discretization instead of spacing-increasing scheme for simplicity. However, the output depth value from ordinal regression is discrete. To obtain the continuous depth from ordinal depth regression, we propose a conditional depth residual estimation layer. The depth residual layer takes image feature $F$ and coarse depth estimation $D$ as input, and infer the residual $D^{'}$ between ordinal depth regression and ground-truth: 
\begin{equation}
\label{eq:depth_residual}
D^{'}=Convs(D,F)
\end{equation}
Hence, combined with the average of pixelwise ordinal loss $\mathcal{L}_{ordinal}$ defined in \cite{fu2018deep} over the entire image domain, our final loss function for monocular depth estimation is: 
\begin{equation}
\label{eq:depth_loss}
\mathcal{L}_{depth}=\mathcal{L}_{ordinal} + ||D+D^{'}-D^*||
\end{equation}
where $D^*$ stands for continuous ground-truth depth map.

\subsection{Multi-view Points Scattering} \label{sec:points_scattering}

\paragraph{Keypoints Generation.} \label{para:keypoints_generation}
Given the camera intrinsics, each predicted depth frame can be back-projected into 3D point cloud. To keep the number of points tractable with varying number of input frames, we need to discard duplicate points that are too close with existing points. Here, three strategies are adopted:
\begin{enumerate}
    \item Only pixels within 2D detection boxes will be converted to point cloud. To keep the 3D distance between sampled points roughly consistent, we compute pixel sampling stride using $stride=\frac{f*r}{d}$, where $f$, $d$, $r$ denote focal length, median value of depth inside 2D box and predefined minimum 3D distance between points respectively.
    \item Before adding points from a new frame, KNN algorithm is utilized to filter out new points that are too close to any existing point.
    \item After concatenating points from all frames, a random sampling is used to limit the maximum number of scene points.
\end{enumerate}

It should be mentioned that, with the help of predicted depth map, the strategy of generating keypoints is flexible. For instance, we can sample several hypothesis points along the depth direction for each pixels. However, this will increase the number of keypoints by several times. Considering that back-projection from multiple frame is approximately a random sampling around object surface, we therefore do not sample hypothesis points along the pixel ray.

\paragraph{Multi-view Feature Aggregation} \label{para:mv_feat_agg}

Inspired by \cite{chen2019point}, we augment the scattered points with multi-view image features. For a 3D point $X$, its projections to $N$ frames are noted as $P=[p_{0},p_{1},...,p_{N-1}]\in R^{N \times 2}$. Feature sampling method such as bilinear interpolation can be used to take the corresponding feature for $P$, which is $F=[f_{0},f_{1},...,f_{N-1}] \in R^{N \times C}$. We define an aggregation function as $f=\alpha(F,M) \in R^{C}$, where $M$ is the mask indicating valid projection and $\eta(\cdot)$ counts the number of \textit{True}. Two aggregation functions are used in our method, 
\begin{enumerate}
    \item \textbf{Mean.} $f=\Bar{f}=\frac{1}{\eta(M)}\sum_{0}^{N-1}(f_i \cdot M_i)$
    \item \textbf{Variance.} $f=\frac{1}{\eta(M)}\sum_{0}^{N-1} M_i \cdot (f_i-\Bar{f})^2$
\end{enumerate}
Note that, essentially, this procedure share the same spirit with \textit{Cost Volume} used by Deep Multi-view Stereo \cite{yao2018mvsnet}. However, it differs in two aspects: (1) The anchor points are learnable and dynamic as they are generated by our Multi-view Points Scattering module, (2) There is no reference frame as each single frame can only observes part of the scene.

Following \cite{vora2020pointpainting}, we append one-hot categorical feature to feature map when 2D detection is available. Though instance mask can provide more accurate semantic information, 2D bounding boxes are used in our method for simplicity.

\begin{figure}[h]
    \centering
    \includegraphics[width=\columnwidth]{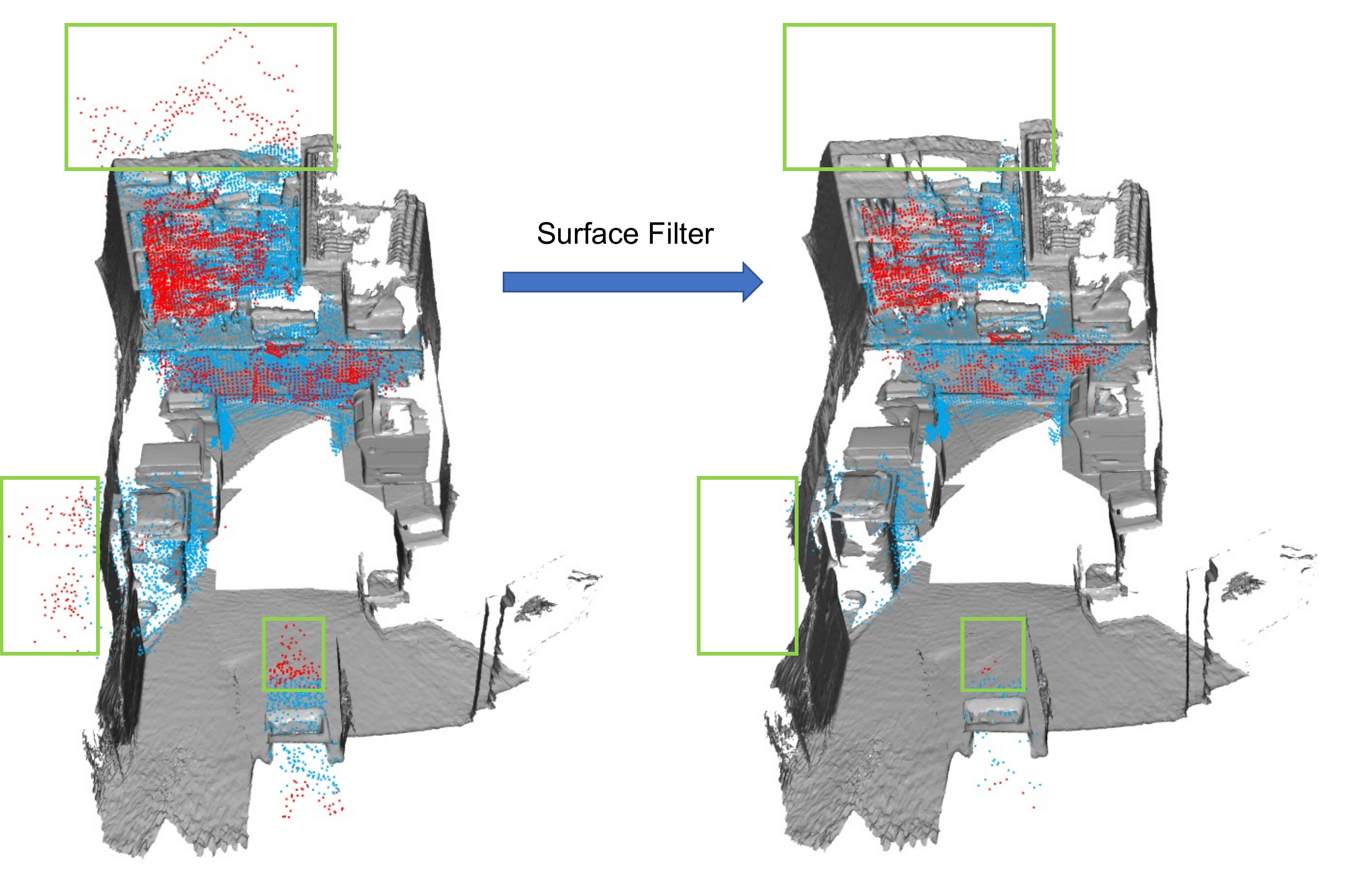}
    \caption{Amount of outliers w/o surface filter. Outliers far away from any object surface will be marked in red, while blue points represent inliers. It is shown that surface filter module helps decrease the number of outliers. }
    \label{fig:surface_filter}
\end{figure}

\paragraph{Surface Filtering.} \label{para:surface_filter}

From section \ref{sec:points_scattering}, a skeleton of the scene is reconstructed, containing some noisy outliers though. Here, outliers refer to those points far away from any object surface by a predefined threshold as shown in Figure \ref{fig:surface_filter}. In this section, we will discuss how to alleviate impact of these outliers to get a cleaner point cloud representation for downstream tasks. Again, the basic idea stems from multi-view stereo method, which assumes that if a pixel is warpped by correct depth to another frame, the appearance of these two corresponding pixels should be similar. Based on this assumption, multi-view feature variance of a point is a suitable criterion for outliers classification. 

In practice, we first lift the image to feature space with three standalone convolution layers, and then aggregate mean and variance of multi-view features for each point. The image feature used by surface filter is designed to be independent of the main 2D backbone, so as to prevent interfering the training of depth estimation. The obtained point cloud features are fed into several Edge Conv layers followed by MLP layers to predict confidence score. As indicated in Figure \ref{fig:overview}, instead of completely eliminating outliers by a predefined threshold, aggregated feature of the scattered points are weighted by the predicted surface score in a soft way. Intuitively, noisy points that violates visual consistency will be down-weighted, making downstream tasks easier. During training, KNN search between the scattered points and ground-truth point cloud is used for label assignment. We use binary focal loss for pointwise classification. Let $X$ denotes the scatter points, $Y$ denotes ground-truth scene points, $p$ denote the estimated score for point $x\in X$.

\begin{equation}
\label{eq:surface_loss}
\begin{aligned}
\mathcal{L}_{surface}&= -(1-p_t)^{\gamma}log(p_t), \\
where, p_t&=
\begin{cases}
    p,   & \text{if } \mathop{min}\limits_{y\in Y}||x-y|| < \tau \\
    1-p, & \text{otherwise}
\end{cases}
\end{aligned}
\end{equation}

\begin{figure*}[htbp]
    \centering
    \includegraphics[width=\textwidth]{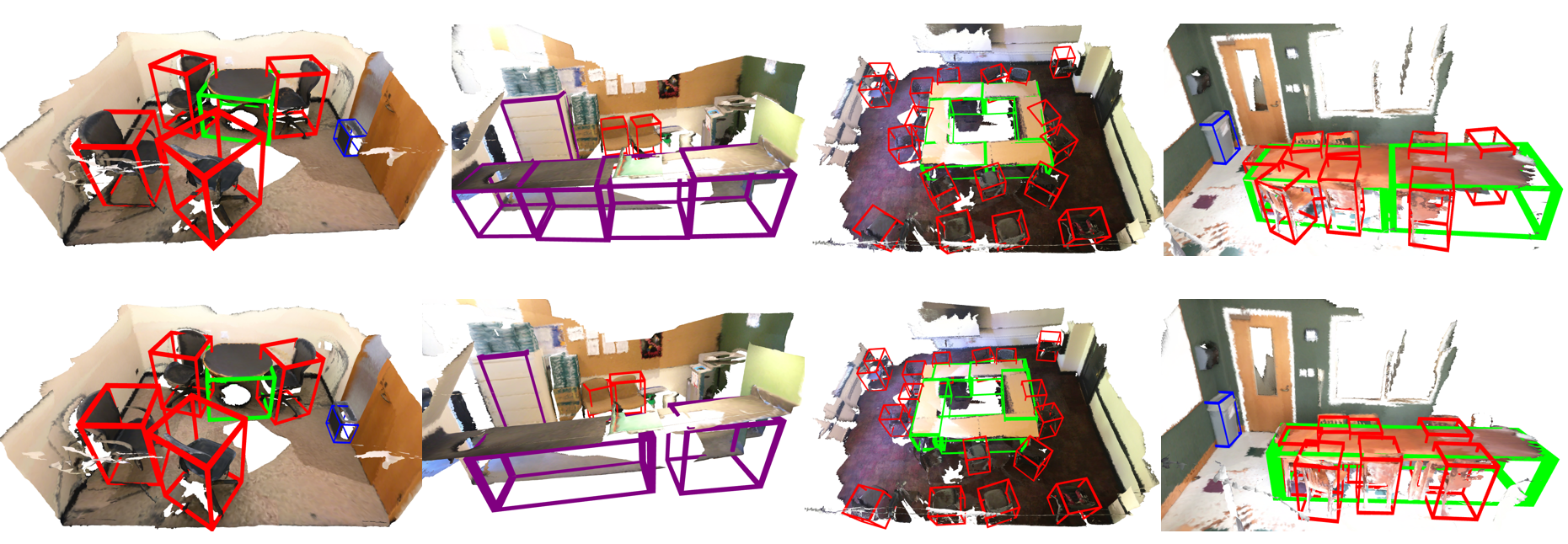}
    \caption{Qualitative result of 3D detection on ScanNet. First row: Ground-truth annotations from Scan2CAD. Second row: Our prediction results for 3D detection. Colors for categories: blue $\rightarrow$ garbage bin, red $\rightarrow$ chair, green $\rightarrow$ table, purple $\rightarrow$ cabinet.}
    \label{fig:qualitative_detect}
\end{figure*}

\subsection{3D Object Detection and Reconstruction}

In this section, we propose to build a 3D detector on top of the final multi-view features augmented points. Different from traditional point cloud 3D detection, the pseudo point cloud is generated by previous stages and associated with multi-view features. First, to extract 3D features, we voxelize the augmented scattered points to feed them into a sparse convolution U-Net backbone. Then, a VoteNet-based 3D detector is applied to infer 3D object oriented bounding boxes in the scene. Following convention, 3D bounding box is represented as $[x,y,z,w,h,d,r_z]$. $[x,y,z]$ is the offset from proposal center to ground-truth object center, $[w,h,d]$ indicates object size, $r_z$ is the rotation around $z$-axis (assuming zero roll and pitch). In addition, object category is predicted by another classification head. The detection losses are the same as in \cite{qi2019deep}.

As a bonus, a reconstruction head can be added in parallel with the detection heads. The reconstruction head shares the same center proposal feature. In order to (1) decouple object pose from shape and (2) get complete object mesh with partial observation, the reconstruction head is only required to estimate a shape code of a pretrained 3D mesh decoder. Here we choose \cite{pan2019deep} pretrained on ShapeNet\cite{chang2015shapenet} as the mesh decoder. The ground-truth shape code is generated offline by the mesh decoder for every instance in our training data, which helps the reconstruction network learn faster. During training, the discrepancy between GT shape code and prediction is minimized. Besides the shape code loss $\mathcal{L}_{code}=||\hat{c}-c||^2$, chamfer distance between target and prediction as well as several regularization terms are also considered. Thus, the overall reconstruction loss is defined as: $\mathcal{L}_{recon}= \mathcal{L}_{code}+ \mathcal{L}_{error} + \mathcal{L}_{CD}+\mathcal{L}_{edge}+ \mathcal{L}_{normal}+ \mathcal{L}_{smoothness}$. For the detailed definition, please refer to the original paper\cite{pan2019deep}. 

\section{Experiments}

\subsection{Dataset and Metrics}
We conduct experiments on ScanNet(V2) \cite{dai2017scannet} with Scan2CAD \cite{avetisyan2019scan2cad} annotations to demonstrate effectiveness of the proposed method. ScanNet is an indoor dataset with ground-truth image poses, mesh reconstruction and 3D instance segmentation labels, however no 3D box annotation is available. Following \cite{runz2020frodo}, the orientated box annotations in \cite{avetisyan2019scan2cad} are used in our experiments. Other data such as camera intrinsic and extrinsic remain the same. We follow the standard training/validation splits(1201 and 312 scans respectively) provided by the benchmark. 

Mean Average Precision (mAP) is commonly used for evaluating 3D object detection. mAP is computed by averaging AP(Average Precision) through semantic classes. AP of each semantic class is adopted as our evaluation metrics. Recall of each class will also be reported.

Chamfer Distance and F-Score is applied as reconstruction metrics. $\mathcal{G}$ and $\mathcal{R}$ denote ground-truth point set and reconstructed point set. Chamfer Distance is formulated as,

\begin{equation}
\label{eq:chamfer_dist}
\begin{aligned}
\mathcal{L}_{CD} = \frac{1}{N_{\mathcal{G}}}\mathop{min}\limits_{p\in\mathcal{G}}||p-q||^2 + \frac{1}{N_{\mathcal{R}}}\mathop{min}\limits_{q\in\mathcal{R}}||p-q||^2
\end{aligned}
\end{equation}

And F-Score is defined as,

\begin{equation}
\label{eq:f_score}
\begin{aligned}
F\mbox{-}Score(d)=\frac{2P(d)R(d)}{P(d)+R(d)} \\
P(d) = \frac{1}{N_{\mathcal{R}}} \sum\limits_{q\in\mathcal{R}} [\mathop{min}\limits_{p\in\mathcal{G}}||p-q||^2<d] \\
R(d) = \frac{1}{N_{\mathcal{G}}} \sum\limits_{p\in\mathcal{G}} [\mathop{min}\limits_{q\in\mathcal{R}}||p-q||^2<d]
\end{aligned}
\end{equation}

\subsection{Implementation Details}
Although our method can be trained end-to-end, we observed that it takes longer time to converge due to the mutual adaptation of different modules. Therefore, the training is divided into three stages: (1) Train Monocular Depth Estimation; (2) Train Depth Estimation and Surface Filter Module from pretrained model of stage-1; (3) Train the whole network with pretrained weights from stage-2. As it is redundant to feed all frames to the network, keyframes are chosen based on a strategy that prefers frames with 2D detection and adequate ego-motion. The number of keyframes are 32 and 50 for training and validation respectively. Adam optimizer with learning rate $5\times 10^{-4}$ is used for all experiments.

\subsection{Evaluation Results}

\begin{table}[h]
\resizebox{\columnwidth}{15mm}{
\begin{tabular}{c||c|c|c|c|c|c}
\hline
\toprule
\multirow{2}{*}{Method}    & \multicolumn{2}{c|}{Chair AP} & \multicolumn{2}{c|}{Table AP} & \multicolumn{2}{c}{Monitor AP}  \\
                           & @0.5          & @0.25         & @0.5         & @0.25      & @0.5          & @0.25 \\ 
\midrule
FroDO                      & 0.32          & -             & 0.06          & -         & 0.04          & - \\ 
MOLTR                      & 0.39          & -             & 0.06          & -         & \textbf{0.10}          & - \\ 
Total3D                    & 0.35          & 0.45          & 0.12          & 0.42      & 0.04          & 0.16 \\
MDR                        & 0.20          & 0.54          & 0.05          & 0.22      & 0.02          & 0.06 \\
Ours                       & \textbf{0.57}          & \textbf{0.81}          & \textbf{0.33}          & \textbf{0.77}      & 0.05          & \textbf{0.30}\\ 
\bottomrule
\end{tabular}
}
\caption{\label{tab:compare_with_others} 3D detection comparison with previous methods on ScanNet-V2 val split. Our methods sample \textbf{50} frames for each scan. Our results reported here are obtained by training a model for each category separately. }
\centering
\end{table}

\begin{table}[h]
\begin{tabularx}{\columnwidth}{CCCCC}
\hline
\toprule
Category                  & AP@0.25       & R@0.25   & AP@0.5        & R@0.5 \\ 
\midrule
Chair                     & 0.76        & 0.89        & 0.51        & 0.62     \\ 
Table                     & 0.67        & 0.83        & 0.29        & 0.41     \\ 
Monitor                   & 0.23        & 0.50        & 0.05        & 0.13     \\ 
Sofa                      & 0.52        & 0.81        & 0.24        & 0.39     \\ 
Bed                       & 0.76        & 0.91        & 0.52        & 0.57     \\ 
Trashbin                & 0.33             & 0.51             & 0.11             & 0.13          \\ 
Bathtub                   & 0.30             & 0.47             & 0.14             & 0.16          \\ 
Bookshelf                 & 0.40             & 0.60             & 0.11             & 0.22          \\ 
Cabinet                   & 0.47             & 0.68             & 0.25             & 0.38          \\ 
mean                      & 0.49             & 0.69             & 0.25             & 0.33          \\ 
\bottomrule
\end{tabularx}
\caption{3D detection result of 9 categories on ScanNet-V2 val split. R@{$t$} stands for Recall with IoU threshold=$t$. All 9 categories are trained with a single model.}
\label{tab:all_class}
\centering
\end{table}


\begin{figure}[h]
    \centering
    \includegraphics[width=\columnwidth]{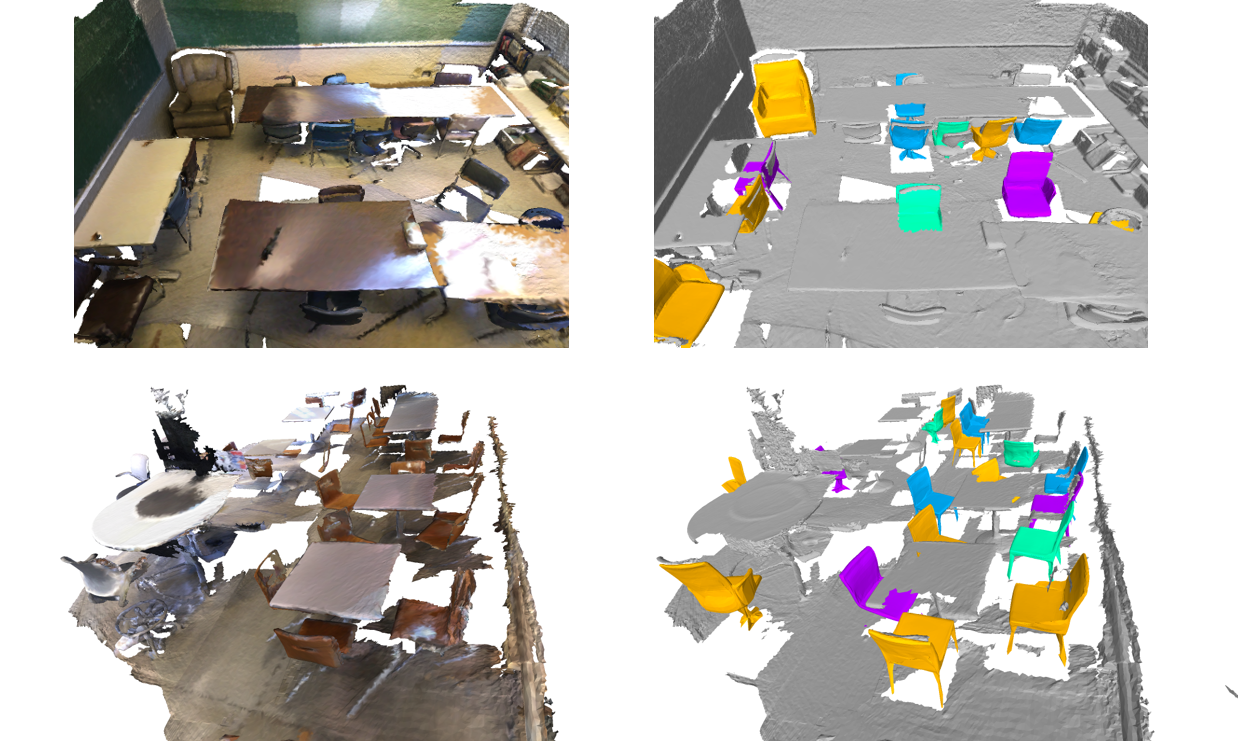}
    \caption{Reconstruction results of chairs on ScanNet. Left: ground-truth mesh scans for reference. Right: reconstruction results. }
    \label{fig:qualitative_recon}
\end{figure}

 As far as we know, \cite{runz2020frodo}\cite{li2021moltr} are the only two multi-frame methods that report metrics of the same Scan2CAD labels. To compare with more other existing methods in the same setting, we extend two single-frame schemes \cite{Nie2020}\cite{Liu2021} to fuse multi-frame predictions. During inference, we estimate 3D detection for every single frame, and then assemble all detection in the same scene coordinate. After that, non-maximum suppression (NMS) is applied with an IoU threshold of 0.01. To align evaluation metrics with FroDO\cite{runz2020frodo} and MOLTR\cite{li2021moltr}, the 11-points AP is calculated in all of our experiments.
 
 As shown in Table \ref{tab:compare_with_others}, our method outperforms FroDO\cite{runz2020frodo} by a margin of 0.24 AP on chair class, without time consuming non-linear optimization used in FroDO. Due to the limitation of single frame FoV, MOLTR\cite{li2021moltr} and other methods that combine 3D detection from each individual frame perform worse than our method, especially those large objects such as tables. Meanwhile, we train on all of 9 categories with a single model, and evaluation results are shown in Table \ref{tab:all_class}.

To measure reconstruction performance of our pipeline, (1) only those boxes that has IoU$\textgreater 0.25$ with any ground-truth box are kept. (2) 2048 points are randomly sampled from both predicted and ground-truth mesh. The ground-truth meshes are taken from \cite{pan2019deep}, where only chair category is available. For F-score, distance threshold is set to 0.004. The results are shown in Table \ref{tab:recon_result}. From \ref{tab:recon_result} we find that when we train multiple categories within a single model, the corresponding AP score for most categories are slightly lower than those in Table \ref{tab:compare_with_others} where we train a model for each separate category, but still higher than other methods we compare with. 

Some qualitative results for both 3D detection and reconstruction can be found in Figure \ref{fig:qualitative_detect} and \ref{fig:qualitative_recon} respectively. As shown by the pictures, the reconstructed meshes are correctly located and akin to the ground-truth objects.

\subsection{Ablation Studies}

\begin{table}[hbt]
\centering
\begin{tabularx}{\columnwidth}{X|X|X|X}
\hline
\toprule
Point Sampling & Multi-view Feature & Surface Filter & AP      \\
\midrule
GS        & \cmark              & \xmark         & 0.33          \\ 
PS        & \xmark              & \xmark         & 0.34          \\ 
PS        & \cmark              & \xmark         & 0.56          \\ 
PS        & \cmark              & \cmark         & 0.57           \\ 
\bottomrule
\end{tabularx}
\caption{\label{tab:ab_test} Ablation study. We compare the metrics of \textit{Chair} with/without key components of our method. \textbf{GS} means \textbf{G}rid \textbf{S}ampling, while \textbf{PS} refers to \textbf{P}oint \textbf{S}cattering}
\end{table}

\begin{table}[hbt]
\begin{tabularx}{\columnwidth}{CCC}
\hline
\toprule
Category & CD & F-score \\ 
\midrule
chair & 0.0100 & 74.33 \\
\bottomrule
\end{tabularx}
\caption{Reconstruction result on ScanNet-V2 val split.}
\label{tab:recon_result}
\end{table}

In order to validate the effectiveness of components in our method, we do several ablation experiments on ScanNet. All reported metrics are evaluated on \textit{chair} class with AP@IoU=0.5.

\paragraph{Dense Grid vs. Scattered Points}
We compare two different point sampling method by keeping other component as much as possible. Nevertheless, three changes are made for $GS$. (1) As using dense grid lead to significantly larger memory footprint, we have to increase the voxel size from 0.04m to 0.16m. (2) Obviously, without multi-view feature, regular grid points carry no information for 3D detection. Therefore, multi-view feature is kept when using $GS$ sampling. (3) We empirically found that $GS$ needs much more proposals due to the Farthest Point Sampling in VoteNet' proposal network. When using same number of proposals(1024) as $PS$, $GS$ only obtains 0.16 AP. Thus, number of proposals is increased to 8192.
From the first two row of Table \ref{tab:ab_test}, using \textit{PS} slightly outperforms dense \textit{GS} even with \textbf{no} multi-view feature and less proposals. By using multi-view feature, $PS$ achieves much better performance than $GS$. 

\paragraph{Multi-view Feature augmentation.}
As shown in Table \ref{tab:ab_test}, naively using back-projected points without augmenting them with multi-view feature deteriorates the performance significantly. The experimental result indicates that the multi-view feature contains rich semantic information.

\paragraph{Surface Filter.}
Although multi-view feature augmented points may capture multi-view geometry implicitly, by modeling it explicitly leads to further improvement for 3D detection task. Surface filter module down-weights those points that go against multi-view geometry to reduce noise for subsequent module.

\section{Conclusion}
In this paper, we present an end-to-end method for 3D object detection and reconstruction with posed monocular images. The key novelty lies in the way of constructing the anchor points. By sampling from multi-view depth prediction, the dynamic anchor points are scattered around object surface. Next, these scattered points are further augmented by multi-view feature and a surface filter module to served as input of 3d detector. Empirical result shows that the proposed method achieves superior performance than its counterparts. 

\bibliographystyle{named}
\bibliography{ijcai22}

\begin{thebibliography}{}

\bibitem[\protect\citeauthoryear{Avetisyan \bgroup \em et al.\egroup
  }{2019}]{avetisyan2019scan2cad}
Armen Avetisyan, Manuel Dahnert, Angela Dai, Manolis Savva, Angel~X Chang, and
  Matthias Nie{\ss}ner.
\newblock Scan2cad: Learning cad model alignment in rgb-d scans.
\newblock In {\em Proceedings of the IEEE/CVF Conference on Computer Vision and
  Pattern Recognition}, pages 2614--2623, 2019.

\bibitem[\protect\citeauthoryear{Bozic \bgroup \em et al.\egroup
  }{2021}]{bozic2021transformerfusion}
Aljaz Bozic, Pablo Palafox, Justus Thies, Angela Dai, and Matthias Nie{\ss}ner.
\newblock Transformerfusion: Monocular rgb scene reconstruction using
  transformers.
\newblock {\em Advances in Neural Information Processing Systems}, 34, 2021.

\bibitem[\protect\citeauthoryear{Chang \bgroup \em et al.\egroup
  }{2015}]{chang2015shapenet}
Angel~X Chang, Thomas Funkhouser, Leonidas Guibas, Pat Hanrahan, Qixing Huang,
  Zimo Li, Silvio Savarese, Manolis Savva, Shuran Song, Hao Su, et~al.
\newblock Shapenet: An information-rich 3d model repository.
\newblock {\em arXiv preprint arXiv:1512.03012}, 2015.

\bibitem[\protect\citeauthoryear{Chen \bgroup \em et al.\egroup
  }{2019}]{chen2019point}
Rui Chen, Songfang Han, Jing Xu, and Hao Su.
\newblock Point-based multi-view stereo network.
\newblock In {\em Proceedings of the IEEE/CVF International Conference on
  Computer Vision}, pages 1538--1547, 2019.

\bibitem[\protect\citeauthoryear{Dai \bgroup \em et al.\egroup
  }{2017}]{dai2017scannet}
Angela Dai, Angel~X Chang, Manolis Savva, Maciej Halber, Thomas Funkhouser, and
  Matthias Nie{\ss}ner.
\newblock Scannet: Richly-annotated 3d reconstructions of indoor scenes.
\newblock In {\em Proceedings of the IEEE Conference on Computer Vision and
  Pattern Recognition}, pages 5828--5839, 2017.

\bibitem[\protect\citeauthoryear{Fu \bgroup \em et al.\egroup
  }{2018}]{fu2018deep}
Huan Fu, Mingming Gong, Chaohui Wang, Kayhan Batmanghelich, and Dacheng Tao.
\newblock Deep ordinal regression network for monocular depth estimation.
\newblock In {\em Proceedings of the IEEE Conference on Computer Vision and
  Pattern Recognition}, pages 2002--2011, 2018.

\bibitem[\protect\citeauthoryear{Li \bgroup \em et al.\egroup
  }{2021}]{li2021moltr}
Kejie Li, Hamid Rezatofighi, and Ian Reid.
\newblock Moltr: Multiple object localization, tracking and reconstruction from
  monocular rgb videos.
\newblock {\em IEEE Robotics and Automation Letters}, 6(2):3341--3348, 2021.

\bibitem[\protect\citeauthoryear{Liu and Liu}{2021}]{Liu2021}
Feng Liu and Xiaoming Liu.
\newblock {Voxel-based 3D Detection and Reconstruction of Multiple Objects from
  a Single Image}.
\newblock 2021.

\bibitem[\protect\citeauthoryear{Luiten \bgroup \em et al.\egroup
  }{2019}]{Luiten2019}
Jonathon Luiten, Tobias Fischer, and Bastian Leibe.
\newblock {Track to reconstruct and reconstruct to track}.
\newblock {\em IEEE Robotics and Automation Letters}, pages 1803--1810, 2019.

\bibitem[\protect\citeauthoryear{Murez \bgroup \em et al.\egroup
  }{2020}]{murez2020atlas}
Zak Murez, Tarrence van As, James Bartolozzi, Ayan Sinha, Vijay Badrinarayanan,
  and Andrew Rabinovich.
\newblock Atlas: End-to-end 3d scene reconstruction from posed images.
\newblock {\em arXiv preprint arXiv:2003.10432}, 2020.

\bibitem[\protect\citeauthoryear{Nicholson \bgroup \em et al.\egroup
  }{2018}]{nicholson2018quadricslam}
Lachlan Nicholson, Michael Milford, and Niko S{\"u}nderhauf.
\newblock Quadricslam: Dual quadrics from object detections as landmarks in
  object-oriented slam.
\newblock {\em IEEE Robotics and Automation Letters}, 4(1):1--8, 2018.

\bibitem[\protect\citeauthoryear{Nie \bgroup \em et al.\egroup
  }{2020}]{Nie2020}
Yinyu Nie, Xiaoguang Han, Shihui Guo, Yujian Zheng, Jian Chang, and Jian~Jun
  Zhang.
\newblock {Total3DUnderstanding: Joint Layout, Object Pose and Mesh
  Reconstruction for Indoor Scenes from a Single Image}.
\newblock {\em Proceedings of the IEEE Computer Society Conference on Computer
  Vision and Pattern Recognition}, pages 52--61, 2020.

\bibitem[\protect\citeauthoryear{Pan \bgroup \em et al.\egroup
  }{2019}]{pan2019deep}
Junyi Pan, Xiaoguang Han, Weikai Chen, Jiapeng Tang, and Kui Jia.
\newblock Deep mesh reconstruction from single rgb images via topology
  modification networks.
\newblock In {\em Proceedings of the IEEE/CVF International Conference on
  Computer Vision}, pages 9964--9973, 2019.

\bibitem[\protect\citeauthoryear{Qi \bgroup \em et al.\egroup
  }{2019}]{qi2019deep}
Charles~R Qi, Or~Litany, Kaiming He, and Leonidas~J Guibas.
\newblock Deep hough voting for 3d object detection in point clouds.
\newblock In {\em Proceedings of the IEEE/CVF International Conference on
  Computer Vision}, pages 9277--9286, 2019.

\bibitem[\protect\citeauthoryear{Reading \bgroup \em et al.\egroup
  }{2021}]{Reading2021}
Cody Reading, Ali Harakeh, Julia Chae, and Steven~L. Waslander.
\newblock {Categorical Depth Distribution Network for Monocular 3D Object
  Detection}.
\newblock 2021.

\bibitem[\protect\citeauthoryear{Ren \bgroup \em et al.\egroup
  }{2016}]{ren2016faster}
Shaoqing Ren, Kaiming He, Ross Girshick, and Jian Sun.
\newblock Faster r-cnn: towards real-time object detection with region proposal
  networks.
\newblock {\em IEEE transactions on pattern analysis and machine intelligence},
  39(6):1137--1149, 2016.

\bibitem[\protect\citeauthoryear{Runz \bgroup \em et al.\egroup
  }{2020}]{runz2020frodo}
Martin Runz, Kejie Li, Meng Tang, Lingni Ma, Chen Kong, Tanner Schmidt, Ian
  Reid, Lourdes Agapito, Julian Straub, Steven Lovegrove, et~al.
\newblock Frodo: From detections to 3d objects.
\newblock In {\em Proceedings of the IEEE/CVF Conference on Computer Vision and
  Pattern Recognition}, pages 14720--14729, 2020.

\bibitem[\protect\citeauthoryear{Vora \bgroup \em et al.\egroup
  }{2020}]{vora2020pointpainting}
Sourabh Vora, Alex~H Lang, Bassam Helou, and Oscar Beijbom.
\newblock Pointpainting: Sequential fusion for 3d object detection.
\newblock In {\em Proceedings of the IEEE/CVF conference on computer vision and
  pattern recognition}, pages 4604--4612, 2020.

\bibitem[\protect\citeauthoryear{Yan \bgroup \em et al.\egroup
  }{2018}]{yan2018second}
Yan Yan, Yuxing Mao, and Bo~Li.
\newblock Second: Sparsely embedded convolutional detection.
\newblock {\em Sensors}, 18(10):3337, 2018.

\bibitem[\protect\citeauthoryear{Yao \bgroup \em et al.\egroup
  }{2018}]{yao2018mvsnet}
Yao Yao, Zixin Luo, Shiwei Li, Tian Fang, and Long Quan.
\newblock Mvsnet: Depth inference for unstructured multi-view stereo.
\newblock In {\em Proceedings of the European Conference on Computer Vision
  (ECCV)}, pages 767--783, 2018.

\end{thebibliography}

\end{document}